\documentclass[10pt,twocolumn,letterpaper]{article}

\usepackage{cvpr}
\usepackage{times}
\usepackage{epsfig}
\usepackage{graphicx}
\usepackage{amsmath}
\usepackage{amssymb}

\usepackage[pagebackref=true,breaklinks=true,letterpaper=true,colorlinks,bookmarks=false]{hyperref}

\cvprfinalcopy 

\def\cvprPaperID{2} 

\ifcvprfinal\fi
\begin{document}

\title{Towards Good Practices for Very Deep Two-Stream ConvNets}

\author{Limin Wang$^1$ \quad \quad Yuanjun Xiong$^1$ \quad \quad Zhe Wang$^2$ \quad \quad Yu Qiao$^2$ \\
\small $^1$Department of Information Engineering, The Chinese University of Hong Kong, Hong Kong \\
\small $^2$Shenzhen key lab of Comp. Vis. \& Pat. Rec.,  Shenzhen Institutes of Advanced Technology, CAS, China \\
{\tt\small \{07wanglimin,bitxiong,buptwangzhe2012\}@gmail.com, yu.qiao@siat.ac.cn}
}

\maketitle

\begin{abstract}
  Deep convolutional networks have achieved great success for object recognition in still images. However, for action recognition in videos, the improvement of deep convolutional networks is not so evident. We argue that there are two reasons that could probably explain this result. First the current network architectures (e.g. Two-stream ConvNets \cite{SimonyanZ14}) are relatively shallow compared with those very deep models in image domain (e.g. VGGNet \cite{SimonyanZ14a}, GoogLeNet \cite{SzegedyLJSRAEVR14}), and therefore their modeling capacity is constrained by their depth. Second, probably more importantly, the training dataset of action recognition is extremely small compared with the ImageNet dataset, and thus it will be easy to over-fit on the training dataset.
  
  To address these issues, this report presents very deep two-stream ConvNets for action recognition, by adapting recent very deep architectures into video domain. However, this extension is not easy as the size of action recognition is quite small. We design several good practices for the training of very deep two-stream ConvNets, namely (i) pre-training for both spatial and temporal nets, (ii) smaller learning rates, (iii) more data augmentation techniques, (iv) high drop out ratio. Meanwhile, we extend the Caffe toolbox into Multi-GPU implementation with high computational efficiency and low memory consumption. We verify the performance of very deep two-stream ConvNets on the dataset of UCF101 and it achieves the recognition accuracy of $91.4\%$.
\end{abstract}

\section{Introduction}
Human action recognition has become an important problem in computer vision and received a lot of research interests in this community \cite{SimonyanZ14,WangS13a,WangQT15a}. The problem of action recognition is challenging due to the large intra-class variations, low video resolution, high dimension of video data, and so on.

The past several years have witnessed great progress on action recognition from short clips \cite{Lan15,Ng15,SimonyanZ14,WangS13a,WangQT13b,WangQT13a,WangQT15a}. These research works can be roughly categorized into two types. The first type of algorithm focuses on the \emph{hand-crafted} local features and \emph{Bag of Visual Words} (BoVWs) representation. The most successful example is to extract improved trajectory features \cite{WangS13a} and employ Fisher vector representation \cite{SanchezPMV13}. The second type of algorithm utilizes \emph{deep convolutional networks} (ConvNets) to learn video representation from raw data (e.g. RGB images or optical flow fields) and train recognition system in an end-to-end manner. The most competitive deep model is the two-stream ConvNets \cite{SimonyanZ14}.

However, unlike image classification \cite{KrizhevskySH12}, deep ConvNets did not yield significant improvement over these traditional methods. We argue that there are two possible reasons to explain this phenomenon. First, the concept of action is more complex than object and it is relevant to other high-level vision concepts, such as interacting object, scene context, human pose. Intuitively, the more complicated problem will need the model of higher complexity. However, the current two-stream ConvNets are relatively shallow (5 convolutional layers and 3 fully-connected layers) compared with those successful models in image classification \cite{SimonyanZ14a,SzegedyLJSRAEVR14}. Second, the dataset of action recognition is extremely small compared the ImageNet dataset \cite{DengDSLL009}. For example, the UCF101 dataset \cite{Soomro12} only contains $13,320$ clips. However, these deep ConvNets always require a huge number of training samples to tune the network weights.

In order to address these issues, this report presents very deep two-stream ConvNets for action recognition. Very deep two-stream ConvNets contain high modeling capacity and are capable of handling the large complexity of action classes. However, due to the second problem above, training very deep models in such a small dataset is much challenging due to the over-fitting problem. We propose several good practices to make the training of very deep two-stream ConvNets stable and reduce the effect of over-fitting. By carefully training our proposed very deep ConvNets on the action dataset, we are able to achieve the state-of-the-art performance on the dataset of UCF101. Meanwhile, we extend the Caffe toolbox \cite{JiaSDKLGGD14} into multi-GPU implementation with high efficiency and low memory consumption. 

The remainder of this report is organized as follows. In Section \ref{sec:method}, we introduce our proposed very deep two-stream ConvNets in details, including network architectures, training details, testing strategy. We report our experimental results on the dataset of UCF101 in Section \ref{sec:experiment}. Finally, we conclude our report in Section \ref{sec:conclusion}.

\section{Very Deep Two-stream ConvNets}
\label{sec:method}

In this section, we give a detailed description of our proposed method. We first introduce the architectures of very deep two-stream ConvNets. After that, we present the training details, which are very important to reduce the effect of over-fitting. Finally, we describe our testing strategies for action recognition.

\subsection{Network architectures}
Network architectures are of great importance in the design of deep ConvNets. In the past several years, many famous network structures have been proposed for image classification, such as AlexNet \cite{KrizhevskySH12}, ClarifaiNet \cite{ZeilerF14}, GoogLeNet \cite{SzegedyLJSRAEVR14}, VGGNet \cite{SimonyanZ14a}, and so on. Some trends emerge during the evolution of from AlexNet to VGGNet: smaller convolutional kernel size, smaller convolutional strides, and deeper network architectures. These trends have turned out to be effective on improving object recognition performance. However, their influence on action recognition has not be fully investigated in video domain. Here, we choose two latest successful network structures to design very deep two-stream ConvNets, namely GoogLeNet and VGGNet.

\textbf{GoogLeNet.} It is essentially a deep convolutional network architecture codenamed \emph{Inception}, whose basic idea is Hebbian principle and the intuition of multi-scale processing. An important component in Inception network is the Inception module. Inception module is composed of multiple convolutional filters with different sizes alongside each other. In order to speed up the computational efficiency, $1 \times 1$ convolutional operation is chosen for dimension reduction. GoogLeNet is a 22-layer network consisting of Inception modules stacked upon each other, with occasional max-pooling layers with stride 2 to halve the resolution of grid. More details can be found in its original paper \cite{SzegedyLJSRAEVR14}.

\textbf{VGGNet.} It is a new convolutional architecture with smaller convolutional size ($3 \times 3$), smaller convolutional stride ($1 \times 1$), smaller pooling window ($2 \times 2$), deeper structure (up to 19 layers). The VGGNet systematically investigates the influence of network depth on the recognition performance, by building and pre-training deeper architectures based on the shallower ones. Finally, two successful network structures are proposed for the ImageNet challenge: VGG-16 (13 convolutional layers and 3 fully-connected layers) and VGG-19 (16 convolutional layers and 3 fully-connected layers). More details can be found in its original paper \cite{SimonyanZ14a}.

\textbf{Very Deep Two-stream ConvNets.} Following these successful architectures in object recognition, we adapt them to the design of two-stream ConvNets for action recognition in videos, which we called \emph{very deep two-stream ConvNets}. We empirically study both GoogLeNet and VGG-16 for the design of very deep two-stream ConvNets. The spatial net is built on a single frame image ($224 \times 224 \times 3$) and therefore its architecture is the same as those for object recognition in image domain. The input of temporal net is 10-frame stacking of optical flow fields ($224 \times 224 \times 20$) and thus the convolutional filters in the first layer are different from those of image classification models.

\subsection{Network training}
Here we describe how to train very deep two-stream ConvNets on the UCF101 dataset. The UCF101 dataset contains $13,320$ video clips and provides 3 splits for evaluation. For each split, there are around $10,000$ clips for training and $3300$ clips for testing. As the training dataset is extremely small and the concept of action is relatively complex, training very deep two-stream ConvNets is quite challenging. From our empirical explorations, we discover several good practices for training very deep two-stream ConvNets as follows.

\textbf{Pre-training for Two-stream ConvNets.} Pre-training has turned out to be an effective way to initialize deep ConvNets when there is not enough training samples available. For spatial nets, as in \cite{SimonyanZ14}, we choose the ImageNet models as the initialization for network training. For temporal net, its input modality are optical flow fields, which capture the motion information and are different from static RGB images. Interestingly, we observe that it still works well by pre-training temporal nets with ImageNet model. In order to make this pre-training reasonable, we make several modifications on optical flow fields and ImageNet model. First, we extract optical flow fields for each video and discretize optical flow fields into interval of $[0,255]$ by a linear transformation. Second, as the input channel number for temporal nets is different from that of spatial nets (20 vs. 3), we average the ImageNet model filters of first layer across the channel, and then copy the average results $20$ times as the initialization of temporal nets.

\textbf{Smaller Learning Rate.} As we pre-trained the two-stream ConvNets with ImageNet model, we use a smaller learning rate compared with original training in \cite{SimonyanZ14}. Specifically, we set the learning rate as follows:
\begin{itemize}
  \item For temporal net, the learning rate starts with $0.005$, decreases to its 1/10 every 10,000 iterations, stops at 30,000 iterations.
  \item For spatial net, the learning rate starts with $0.001$, decreases to its 1/10 every 4,000 iterations, stops at 10,000 iterations.
\end{itemize}
In total, the learning rate is decreased 3 times. At the same time, we notice that it requires less iterations for the training of very deep two-stream ConvNets. We analyze that this may be due to the fact we pre-trained the networks with the ImageNet models.

\textbf{More Data Augmentation Techniques.} It has been demonstrated that data augmentation techniques such as random cropping and horizontal flipping are very effective to avoid the problem of over-fitting. Here, we try two new data augmentation techniques for training very deep two-stream ConvNets as follows:
\begin{itemize}
  \item We design a corner cropping strategy, which means we only crop 4 corners and 1 center of the images. We find that if we use random cropping method, it is more likely select the regions close to the image center and training loss goes quickly down, leading to the over-fitting problem. However, if we constrain the cropping to the 4 corners or 1 center explicitly, the variations of input to the network will increase and it helps to reduce the effect of over-fitting.
  \item We use a multi-scale cropping method for training very deep two-stream ConvNets. Multi-scale representations have turned out to be effective for improving the performance of object recognition on the ImageNet dataset \cite{SimonyanZ14a}. Here, we adapt this good practice into the task of action recognition. But we present an efficient implementation compared with that in object recognition \cite{SimonyanZ14a}. We fix the input image size as $256 \times 340$ and randomly sample the cropping width and height from \{256,224,192,168\}. After that, we resize the cropped regions to $224 \times 224$. It is worth noting that this cropping strategy not only introduces the multi-scale augmentation, but also aspect ratio augmentation.
\end{itemize}

\textbf{High Dropout Ratio.} Similar to the original two-stream ConvNets \cite{SimonyanZ14}, we also set high drop out ratio for the fully connected layers in the very deep two-stream ConvNets. In particular, we set 0.9 and 0.8 drop out ratios for the fully connected layers of temporal nets. For spatial nets, we set 0.9 and 0.9 drop out ratios for the fully connected layers .

\textbf{Multi-GPU training.}  One great obstacle for applying deep learning models in video action recognition task is the prohibitively long training time. Also the input of multiple frames heightens the memory consumption for storing layer activations. We solve these problems by employing data-parallel training on multiple GPUs. The training system is implemented with Caffe \cite{JiaSDKLGGD14} and OpenMPI. Following a similar technique used in \cite{HeArxiv2015}, we avoid synchronizing the parameters of fully connected (fc) layers by gathering the activations from all worker processes before running the fc layers. With 4 GPUs, the training is 3.7x faster for VGGNet-16 and 4.0x faster for GoogLeNet. It takes 4x less memory per GPU. The system is publicly available \footnote{\url{https://github.com/yjxiong/caffe/tree/action_recog}}.

\begin{table*}
\begin{center}
\resizebox{1\textwidth}{!}{
\begin{tabular}{|c|c|c|c|c|c|c|c|c|c|c|c|c|}
  \hline
  &\multicolumn{4}{|c|}{Spatial nets} & \multicolumn{4}{c|}{Temporal nets} &  \multicolumn{4}{c|}{Two-stream ConvNets} \\ \hline
  Architecture & Split1 & Split2 & Split3 & Average & Split1 & Split2 & Split3 & Average & Split1 & Split2 & Split3 & Average \\ \hline
  ClarifaiNet (from \cite{SimonyanZ14}) & 72.7\% & - & - & 73.0\% & 81.0\% & - & - &  83.7\% & 87.0\% & - & - &  88.0\% \\
  GoogLeNet & 77.1\% & 73.2\% & 75.6\% & 75.3\% & 83.9\% & 86.5\% & 86.9\% & 85.8\% & 89.0\% & 89.3\% & 89.5\% & 89.3\%\\
  VGGNet-16 & 79.8\% & 77.3\% & 77.8\% & {\bf 78.4\%} & 85.7\% & 88.2\% & 87.4\% & {\bf 87.0\%} & 90.9\% & 91.6\% & 91.6\% & {\bf 91.4\%} \\
  \hline
\end{tabular}
}
\vspace{2mm}
\caption{Performance comparison of different architectures on the UCF101 dataset. ({\bf with} using our proposed good practices)}
\label{tbl:result_ucf}
\end{center}
\end{table*}

\begin{table*}
\begin{center}
\resizebox{0.7\textwidth}{!}{
\begin{tabular}{|c|c|c|c|c|c|}
  \hline
  \multicolumn{3}{|c|}{Spatial nets} & \multicolumn{3}{c|}{Temporal nets}  \\ \hline
   ClarifaiNet & GoogLeNet & VGGNet-16 & ClarifaiNet & GoogLeNet & VGGNet-11 \\ \hline
   42.3\% & 53.7\% & \textbf{54.5\%} & \textbf{47.0\%} & 39.9\% &  42.6\%\\
  \hline
\end{tabular}
}
\vspace{2mm}
\caption{Performance comparison of different architectures on the THUMOS15 \cite{THUMOS15} validation dataset. (from \cite{WangWXQ15}, {\bf without} using our proposed good practices)}
\label{tbl:result_thumos}
\end{center}
\end{table*}

\subsection{Network testing}

For fair comparison with the original two-stream ConvNets \cite{SimonyanZ14}, we follow their testing scheme for action recognition. At the test time, we sample 25 frame images or optical flow fields for the testing of spatial and temporal nets, respectively. From each of these selected frames, we obtain 10 inputs for very deep two-stream ConvNets, i.e. 4 corners, 1 center, and their horizontal flipping. The final prediction score is obtained by averaging across the sampled frames and their cropped regions. For the fusion of spatial and temporal nets, we use a weighted linear combination of their prediction scores, where the weight is set as $2$ for temporal net and $1$ for spatial net.

\section{Experiments}
\label{sec:experiment}

{\bf Datasets and Implementation Details.} In order to verify the effectiveness of proposed very deep two-stream ConvNets, we conduct experiments on the UCF101 \cite{Soomro12} dataset. The UCF101 dataset contains 101 action classes and there are at least 100 video clips for each class. The whole dataset contains $13,320$ video clips, which are divided into 25 groups for each action category. We follow the evaluation scheme of the THUMOS13 challenge \cite{THUMOS13} and adopt the three training/testing splits for evaluation. We report the average recognition accuracy across classes over these three splits. For the extraction of optical flow fields, we follow the work of TDD \cite{WangQT15a} and choose the TVL1 optical flow algorithm \cite{ZachPB07}. Specifically, we use the OpenCV implementation, due to its balance between accuracy and efficiency.

{\bf Results.} We report the action recognition performance in Table \ref{tbl:result_ucf}. We compare three different network architectures, namely ClarifaiNet, GoogLeNet, and VGGNet-16. From these results, we see that the deeper architectures obtains better performance and VGGNet-16 achieves the best performance. For spatial nets, VGGNet-16 outperform shallow network by around $5\%$, and for temporal net, VGGNet-16 is better by around $4\%$. Very deep two-stream ConvNets outperform original two-stream ConvNets by $3.4\%$.

It is worth noting in our previous experience \cite{WangWXQ15} in THUMOS15 Action Recognition Challenge \cite{THUMOS15}, we have tried very deep two-stream ConvNets {\bf but temporal nets with deeper structure did not yield good performance}, as shown in Table \ref{tbl:result_thumos}. In this THUMOS15 submission, we train the very deep two-stream ConvNets in the same way as the original two-stream ConvNets \cite{SimonyanZ14} without using these proposed good practices. From the different performance of very deep two-stream ConvNets on two datasets, we conjecture that our proposed good practices is very effective to reduce the effect of over-fitting due to (a) pre-training temporal nets with the ImageNet models; (b) using more data augmentation techniques.

\begin{table}
\begin{center}
\resizebox{0.35\textwidth}{!}{
\begin{tabular}{|c|c|c|}
  \hline
  Method & Year & Accuracy \\
  \hline
  iDT+FV \cite{WangS13a} & 2013 & 85.9\% \\
  iDT+HSV \cite{PengWWQ14} & 2014 & 87.9\% \\
  MIFS+FV \cite{Lan15} & 2015 & 89.1\% \\
  TDD+FV \cite{WangQT15a} & 2015 & 90.3\% \\
  \hline
  DeepNet \cite{KarpathyTSLSF14} & 2014 & 63.3\% \\
  Two-stream \cite{SimonyanZ14} & 2014 & 88.0\% \\
  Two-stream+LSTM \cite{Ng15} & 2015 & 88.6\% \\
  \hline
  Very deep two-stream & 2015 & {\bf 91.4\%} \\
  \hline
\end{tabular}
}
\vspace{2mm}
\caption{Performance comparison with the state of the art on UCF101 dataset.}
\label{tbl:comparison}
\end{center}
\end{table}

{\bf Comparison.} Finally, we compare our recognition accuracy with several recent methods and the results are shown in Table \ref{tbl:comparison}. We first compare with Fisher vector representation of hand-crafted features like Improved Trajectories (iDT) \cite{WangS13a} or deep-learned features like Trajectory-Pooled Deep-Convolutional Descriptors (TDD) \cite{WangQT15a}. Our results is better than all these Fisher vector representations. Second, we perform comparison between the very deep two-stream ConvNets with other deep networks such as DeepNets \cite{KarpathyTSLSF14} and two-stream ConvNets with recurrent neural networks \cite{Ng15}. We see that our proposed very deep models outperform previous ones and are better than the best result by $2.8\%$.

\section{Conclusions}
\label{sec:conclusion}

In this work we have evaluated very deep two-stream ConvNets for action recognition. Due to the fact that action recognition dataset is extremely small, we proposed several good practices for the training very deep two-stream ConvNets. With our carefully designed training strategies, the proposed very deep two-stream ConvNets achieved the recognition accuracy of $91.4\%$ on the UCF101 dataset. Meanwhile, we extended the famous Caffe toolbox into Multi-GPU implementation with high efficiency and low memory consumption.

{
\bibliographystyle{ieee}
\bibliography{deep}

\begin{thebibliography}{10}\itemsep=-1pt

\bibitem{DengDSLL009}
J.~Deng, W.~Dong, R.~Socher, L.~Li, K.~Li, and F.~Li.
\newblock {ImageNet}: {A} large-scale hierarchical image database.
\newblock In {\em CVPR}, pages 248--255, 2009.

\bibitem{THUMOS15}
A.~Gorban, H.~Idrees, Y.-G. Jiang, A.~Roshan~Zamir, I.~Laptev, M.~Shah, and
  R.~Sukthankar.
\newblock {THUMOS} challenge: Action recognition with a large number of
  classes.
\newblock \url{http://www.thumos.info/}, 2015.

\bibitem{HeArxiv2015}
K.~He, X.~Zhang, S.~Ren, and J.~Sun.
\newblock Delving deep into rectifiers: Surpassing human-level performance on
  imagenet classification.
\newblock {\em CoRR}, abs/1502.01852, 2015.

\bibitem{JiaSDKLGGD14}
Y.~Jia, E.~Shelhamer, J.~Donahue, S.~Karayev, J.~Long, R.~B. Girshick,
  S.~Guadarrama, and T.~Darrell.
\newblock Caffe: Convolutional architecture for fast feature embedding.
\newblock {\em CoRR}, abs/1408.5093.

\bibitem{THUMOS13}
Y.-G. Jiang, J.~Liu, A.~Roshan~Zamir, I.~Laptev, M.~Piccardi, M.~Shah, and
  R.~Sukthankar.
\newblock {THUMOS} challenge: Action recognition with a large number of
  classes, 2013.

\bibitem{KarpathyTSLSF14}
A.~Karpathy, G.~Toderici, S.~Shetty, T.~Leung, R.~Sukthankar, and L.~Fei{-}Fei.
\newblock Large-scale video classification with convolutional neural networks.
\newblock In {\em CVPR}, pages 1725--1732, 2014.

\bibitem{KrizhevskySH12}
A.~Krizhevsky, I.~Sutskever, and G.~E. Hinton.
\newblock {ImageNet} classification with deep convolutional neural networks.
\newblock In {\em NIPS}, pages 1106--1114, 2012.

\bibitem{Lan15}
Z.~Lan, M.~Lin, X.~Li, A.~G. Hauptmann, and B.~Raj.
\newblock Beyond gaussian pyramid: Multi-skip feature stacking for action
  recognition.
\newblock In {\em CVPR}, pages 204--212, 2015.

\bibitem{Ng15}
J.~Y.-H. Ng, M.~Hausknecht, S.~Vijayanarasimhan, O.~Vinyals, R.~Monga, and
  G.~Toderici.
\newblock Beyond short snippets: Deep networks for video classification.
\newblock In {\em CVPR}, pages 4694--4702, 2015.

\bibitem{PengWWQ14}
X.~Peng, L.~Wang, X.~Wang, and Y.~Qiao.
\newblock Bag of visual words and fusion methods for action recognition:
  Comprehensive study and good practice.
\newblock {\em CoRR}, abs/1405.4506, 2014.

\bibitem{SanchezPMV13}
J.~S{\'a}nchez, F.~Perronnin, T.~Mensink, and J.~J. Verbeek.
\newblock Image classification with the fisher vector: Theory and practice.
\newblock {\em International Journal of Computer Vision}, 105(3):222--245,
  2013.

\bibitem{SimonyanZ14}
K.~Simonyan and A.~Zisserman.
\newblock Two-stream convolutional networks for action recognition in videos.
\newblock In {\em NIPS}, pages 568--576, 2014.

\bibitem{SimonyanZ14a}
K.~Simonyan and A.~Zisserman.
\newblock Very deep convolutional networks for large-scale image recognition.
\newblock {\em CoRR}, abs/1409.1556, 2014.

\bibitem{Soomro12}
K.~Soomro, A.~R. Zamir, and M.~Shah.
\newblock {UCF101}: A dataset of 101 human actions classes from videos in the
  wild.
\newblock {\em CoRR}, abs/1212.0402, 2012.

\bibitem{SzegedyLJSRAEVR14}
C.~Szegedy, W.~Liu, Y.~Jia, P.~Sermanet, S.~Reed, D.~Anguelov, D.~Erhan,
  V.~Vanhoucke, and A.~Rabinovich.
\newblock Going deeper with convolutions.
\newblock {\em CoRR}, abs/1409.4842, 2014.

\bibitem{WangS13a}
H.~Wang and C.~Schmid.
\newblock Action recognition with improved trajectories.
\newblock In {\em ICCV}, pages 3551--3558, 2013.

\bibitem{WangQT13b}
L.~Wang, Y.~Qiao, and X.~Tang.
\newblock Mining motion atoms and phrases for complex action recognition.
\newblock In {\em ICCV}, pages 2680--2687, 2013.

\bibitem{WangQT13a}
L.~Wang, Y.~Qiao, and X.~Tang.
\newblock Motionlets: Mid-level 3{D} parts for human motion recognition.
\newblock In {\em CVPR}, pages 2674--2681, 2013.

\bibitem{WangQT15a}
L.~Wang, Y.~Qiao, and X.~Tang.
\newblock Action recognition with trajectory-pooled deep-convolutional
  descriptors.
\newblock In {\em CVPR}, pages 4305--4314, 2015.

\bibitem{WangWXQ15}
L.~Wang, Z.~Wang, Y.~Xiong, and Y.~Qiao.
\newblock {CUHK\&SIAT} submission for {THUMOS15} action recognition challenge.
\newblock In {\em {THUMOS'15 Action Recognition Challenge}}, 2015.

\bibitem{ZachPB07}
C.~Zach, T.~Pock, and H.~Bischof.
\newblock A duality based approach for realtime tv-{$L^1$} optical flow.
\newblock In {\em 29th {DAGM} Symposium on Pattern Recognition}, 2007.

\bibitem{ZeilerF14}
M.~D. Zeiler and R.~Fergus.
\newblock Visualizing and understanding convolutional networks.
\newblock In {\em ECCV}, pages 818--833, 2014.

\end{thebibliography}
}

\end{document}